\documentclass[letterpaper, 10 pt, conference]{ieeeconf}
\IEEEoverridecommandlockouts                              
\usepackage{cite}
\usepackage{amsmath,amssymb,amsfonts}
\usepackage{algorithmic}
\usepackage{graphicx}
\usepackage{textcomp}
\usepackage{xcolor}
\usepackage{hyperref}

\title{A Small Claims Court for the NLP: Judging Legal Text Classification Strategies With Small Datasets\\

}

\author{Mariana Noguti$^{1}$, Eduardo Vellasques$^{2}$ and Luiz S. Oliveira$^1$
\thanks{This work has been supported by the Brazilian National Council for Scientific and Technological Development (CNPq) -- Grant 303298/2022-7}
\thanks{$^{1}$Mariana Noguti (mynoguti@mppr.mp.br) and Luiz S. Oliveira are with the Department of Informatics of Universidade Federal do Paraná, Curitiba, Brazil. Mariana Noguti is also with the Ministério Público do Paraná, Curitiba, Brazil.}%
\thanks{$^{2}$Eduardo Vellasques is with SAP SE, Germany}%
}

\begin{document}

\maketitle

\begin{abstract}
Recent advances in language modelling has significantly decreased the need of labelled data in text classification tasks. Transformer-based models, pre-trained on unlabeled data, can outmatch the performance of models trained from scratch for each task. However, the amount of labelled data need to fine-tune such type of model is still considerably high for domains requiring expert-level annotators, like the legal domain. This paper investigates the best strategies for optimizing the use of a small labeled dataset and large amounts of unlabeled data and perform a classification task in the legal area with 50 predefined topics. More specifically, we use the records of demands to a Brazilian Public Prosecutor's Office aiming to assign the descriptions in one of the subjects, which currently demands deep legal knowledge for manual filling. The task of optimizing the performance of classifiers in this scenario is especially challenging, given the low amount of resources available regarding the Portuguese language, especially in the legal domain. Our results demonstrate that classic supervised models such as logistic regression and SVM and the ensembles random forest and gradient boosting achieve better performance along with embeddings extracted with word2vec when compared to BERT language model. The latter demonstrates superior performance in association with the architecture of the model itself as a classifier, having surpassed all previous models in that regard. The best result was obtained with Unsupervised Data Augmentation (UDA), which jointly uses BERT, data augmentation, and strategies of semi-supervised learning, with an accuracy of 80.7\% in the aforementioned task.

\end{abstract}

\section{Introduction}

Text classification is a widely used task in Natural Language Processing (NLP) which consists of predicting labels associated with textual content. To obtain models capable of performing such complex tasks, a significant amount of labeled examples is essential to extract a representative sample of each class and distinguish them. This requirement has a propensity to grow exponentially with the increase in the number of classes to predict and the variability of the vocabulary, making a project difficult or even unfeasible due to the costs involved in the collection of suitable material.

Some of the main approaches towards the optimization of limited labeled corpora are the use of transfer learning and data augmentation (DA). The first has consistently improved the performance of several NLP tasks while decreasing the need for large amounts of labeled data. DA is a well-established technique in the field of image classification, with increased interest in NLP as well, and consists of synthesizing annotated corpora in scarce supervised scenarios. Specifically addressing our research, we intend to explore the aforementioned techniques and develop a suitable classifier with little data. Our dataset was obtained from the records of demands to the Public Prosecutor's Office of the State of Parana (\textit{Ministério Público do Estado do Paraná} - MPPR), Brazil, and consists of brief reports in Portuguese, a low resource language in NLP, especially in the legal domain.

The main duties of MPPR are, among others, to defend the democratic regime and the unavailable individual rights. It has about 700 units distributed throughout the state of Parana, meeting the demands of the population in several areas such as the environment, public health, criminal matters, and many others. Currently, the institution receives thousands of petitions of the population, all registered in an electronic system and validated by a Prosecutor.

To build our dataset, a specialized MPPR team carried out a study to verify whether the legal area of the demands was correctly classified in the system and analyzed 9,387 demands, having verified an error rate of 46.73\% due to the wrong association of the theme in relation to its description. Such analysis has become worrying, given that incorrect registration can lead to the process being forwarded to wrong sectors, increasing its processing time, in addition to the fact that records of demands in legal matters help in decision-making and the allocation of MPPR's resources.

Intending to improve the information quality, we proposed to automate the classification of the demands based on their descriptions. The automation of this task would standardize the classification of demands in the system, generating more reliable statistics and aiding in decision making. Furthermore, registration time would be reduced, removing the need for an expert to validate. To avoid the excessive allocation of human resources in the construction of a reliable dataset, we investigated techniques that potentially improve text classification with few labeled examples. Our best result reached an accuracy of 80.7\% in the prediction of 50 classes, greatly outperforming the hit rate of manual filling.


\section{Related Work}
\label{sec:related}

In Brazil, the high rates of judicialization can be translated through a report from the National Justice Council \cite{CNJ-2020} which stated that, in December 2019, there were more than 77 million legal proceedings pending completion. This fact highlights the importance of automated analysis of legal texts, reason why we bring the important research done in this field.

Sulea et al. \cite{Sulea-2017} used court decisions of the Supreme Court of France to predict several aspects using ensemble systems of SVM classifiers combined with simple representations such as unigrams and bigrams. Undavia et al. \cite{Undavia-2018} proposed to classify US Supreme Court documents through various combinations between feature representation and classification models. Research conducted by Silva et al. \cite{Silva-2018} with Portuguese corpora obtained good performance in classifying documents received by the Supreme Court of Brazil using CNN with a simple embedding layer. Another attempt for the Portuguese language was reported by Aguiar et al. \cite{Aguiar-2021} where the authors discuss the classification of legal documents extracted from lawsuits in Brazilian courts.

Some studies explored more recent language models such as \cite{Soh-2019} where the authores compared embedding techniques for multilabel classification, with the transfer learning approaches performing well in low regime data for BERT base and large, and ULMFit. Clavie et al. \cite{Clavie-2021} compared the BERT model against ULMFit, both trained with legal corpora. They reported good results using ULMFit in short texts scenarios, but for larger texts, the authors experienced a reduced performance of the LSTM-based language model, probably due to its difficulty in retaining long chunks of information, inherent to the recurrent structure.

Elwany et al. \cite{Elwany-2019} used legal texts to fine-tune BERT and perform a legal agreements classification, with an improved performance using fine-tuned BERT. In the same line of research, Tai et al. \cite{Tai-2019} fine-tuned German BERT to the legal vocabulary, finding no relevant differences between the use of generic German BERT and the fine-tuned version in any of the proposed challenges. Chalkidis et al. \cite{Chalkidis-2020} performed similar tests in English, with better results using adapted legal vocabulary, especially in tasks that require in-domain knowledge. The paper of Clavie et al. \cite{Clavie-2021b} compared a baseline model with both generic and specialized BERT for classification. Although there were great improvements using BERT, there was not much performance increase using the specialized BERT when compared to its generic version. The result goes in the opposite direction to other experiments conducted with the specialization of language models, for example with scientific vocabulary \cite{Beltagy-2019} or fine-tuned with biomedical corpora \cite{Lee-2019}. As a hypothesis for this phenomenon, the authors supposed that language models may not yet be able to capture the depth of legal vocabulary.

As the legal domain still has a certain resistance to the use of new technologies and research in the Portuguese language, in particular, is still very incipient, we believe that our research can contribute to bring new insights into this minimally explored area. To our knowledge, this is the first work in Portuguese to explore NLP techniques in the legal domain aimed at optimizing resources in scenarios of data scarcity and limited human and computational resources.

\section{Methodology}
\label{sec:methodology}

The proposed method for this research can be subdivided into three main parts: preprocessing, feature extraction and training the classifiers. Below are the detailed processes of each of these steps.

\subsection{Preprocessing}

Two different preprocessing routines were used, related to word2vec \cite{Mikolov-2013} and BERT \cite{Devlin-2018}. In both cases, we applied the same preprocessing steps used to generate the original language models, to guarantee that our tokens corresponded to the existing tokens in the initial vocabulary and were correctly identified.

For word2vec, simplified preprocessing techniques were used, narrowed to word tokenization, lowercasing, replacing URL and email structures into the tokens ``URL'' and ``EMAIL'' and standardization of numerals to zero.

The BERT model was used with the transformers\footnote{\href{https://huggingface.co/docs/transformers/index}{https://huggingface.co/docs/transformers/index}} architecture, which has the preprocessing routines stored in a tokenizer file. Therefore, we downloaded the Portuguese BERT tokenizer (CITAR O NOME DO TOKENIZER AQUI) and passed our corpora to it, obtaining the processed tokens.

\subsubsection{Data Augmentation}

In addition to the data cleaning processes for feature extraction, we performed two data augmentation (DA) operations suggested in \cite{Xie-2019}. For both cases, exactly one synthetic example was created from each original example in our dataset, thus doubling the number of examples for training.

\textbf{TF-IDF Replacement}: given that some words may be more relevant than others to distinguish classes in text classification, the TF-IDF of each word in our vocabulary was calculated, creating augmented examples from the exchange of words that presented a low relevance for another equally uninformative about the class, in order to prevent changing the ground-truth label of the examples.

\textbf{Back Translation (BT)}: this technique was performed with the use of the Google Translation API, which relies on the Neural Machine Translation model \cite{Wu-2016}. We converted our sentences to English and then back to Portuguese.

\subsection{Feature Extraction}
\label{sec:feature_extraction_methodology}

We used word2vec and BERT language models, both with a pretrained version and a fine-tuned variant, trained with the Internal Procedures Dataset. The four models are described below.

\textbf{Generic word2vec}: Hartmann et al. \cite{Hartmann-2017} trained several language models for their research using Portuguese multigenre corpora and made the resulting embeddings available for download\footnote{\href{http://nilc.icmc.usp.br/embeddings}{http://nilc.icmc.usp.br/ embeddings}}. We chose the word2vec model with Skip-Gram and vector size of 600 as a baseline method for the proximity of the parameters with specialized word2vec.

\textbf{Specialized word2vec}: we trained a specialized word2vec model from scratch with the Skip-Gram task, minimum word count equals to 5, window value of 10 and vector size of 600. These parameters were chosen due to their proximity with the results of \cite{Noguti-2020} that handled a dataset similar to ours.

\textbf{Generic BERT}: Souza et al. \cite{Souza-2019} applied BERT architecture to train a model on more than 3 million documents available on the brWaC corpus \cite{Wagner-2018} with 30k subword units vocabulary. We relied on the BERT-base version\footnote{\href{https://github.com/neuralmind-ai/portuguese-bert}{https://github.com/neuralmind-ai/portuguese-bert}}, which generates vectors of size 768, and limited the sequence length to 128 due to GPU limitations.

\textbf{Specialized BERT}: we fine-tuned BERT-base model to the legal vocabulary, using the script available in the transformers repository\footnote{\href{https://github.com/huggingface/transformers/tree/main/examples/tensorflow/language-modeling}{https://github.com/huggingface/transformers/tree/main/ examples/tensorflow/language-modeling}}. Again, due to computational limitations, our hyperparameter search focused on the learning rate, fixing other parameters and the sequence length to 128.

\subsection{Classifiers}

We initially trained our supervised models with the training data and later retrained them with augmented examples to compare the performance in the presence of larger amounts of data, half of which were synthetic. 

\subsubsection{Supervised Learning}

The four embeddings were applied to the Logistic Regression (LR) and Support Vector Machines (SVM) classifiers, and the ensembles Random Forest (RF) and Gradient Boosting (GB). We chose the LR model due to the fact that it is easy to train and not very dependent on parameter optimizations while SVM was selected for being a popular classifier used in NLP. We also compared two ensemble methods, both relying on decision trees structures, to analyze their performance in our data. 

For optimization, we performed a grid search on the hyperparameters with 5-fold cross validation. Additionally, we tested the use of BT as a data augmentation technique in our training set and retrained the models with the same cross validation methodology. The test set was not modified during these experiments. Specifically with BERT embeddings, we also carried out a classification task by incorporating one additional output layer to its architecture and fine-tuned with our labeled data, but did not experiment augmentation methods.

\subsubsection{Unsupervised Data Augmentation}

Unsupervised Data Augmentation (UDA) is a semi-supervised method introduced by \cite{Xie-2019} that combines labeled and unlabeled data to create robust classification models. The rationale is to create a classifier using a small portion of labeled data and apply the same classifier in each example of the unlabeled data and its augmented version, trying to predict the same label on the original and the synthetic example. It also relies on a novel technique called Training Signal Annealing (TSA), which is a way to control the examples that are passed from the unlabeled data to the labeled data to prevent overfitting.

Taking into account the need for augmented data to perform UDA, we used the TF-IDF replacement and BT data augmentation strategies with the BERT embeddings. We investigated variations in the training and evaluation batch sizes, and the TSA strategies, but could not perform a hyperparameter optimization concerning sentence size or learning rate due to scarce computational resources.

\subsection{Evaluation Metrics}

Since the training, validation, and testing sets were perfectly balanced, we chose the accuracy as the main evaluation metric in our experiments. Additionally, we calculated the precision, recall, and F1-score per class and for the classifier as a whole as a complementary form of analysis.

\section{Experiments}
\label{sec:experiments}

Our research relied on several datasets obtained from MPPR's demands management system. We describe the methodology used to extract and prepare the data in this section. In the following sections, we outline the preprocessing, feature extraction and classification experiments.


\subsection{Datasets}

Initially, we obtained the demands to MPPR between the years 2016 and 2020, containing 492,312 observations and described as Demands Dataset. In face of a reduced number of examples in some legal areas, we selected 50 of the 164 available classes to analyze in this research. To obtain a reliable dataset, a specialized team carried out a process of data audition, manually analyzing and validating the classifications filled in the system and discarding the records that eventually presented the wrong label. Considering the short time available, only 130 observations from each of the 50 themes were validated, totaling 6,500 observations. During the experiments, 46.73\% of the records were discarded due to misclassification, reinforcing the need to automate the task for obtaining more reliable and standardized information. The details of the 50 classes and their validation statistics can be found in the supplementary materials.

After the completion of the review process, the Supervised Dataset was randomized and divided into three partitions for train, validation and test with 3,500, 1,500 and 1,500 examples, respectively. The final training set presented 70 examples of each class while the validation and test set contained 30 examples per class each, all perfectly balanced.



We also obtained 784,358 records corresponding to MPPR registered procedures, including a field with a short text containing the description of the subject in the document, to build up the Internal Procedures Dataset. Due to the fact that these descriptions have a similar vocabulary to the demands, we deemed appropriate their use to train and fine-tune language models to a legal specialized version.

\subsection{Feature Extraction}

We performed four distinct routines to preprocess and embed our texts, related to word2vec and BERT models, both in generic and specialized versions. The word2vec models already had their parameters set. For the generic BERT model, the majority of the parameters were maintained, lacking only a strategy for feature extraction.

Since BERT has an architecture with twelve identical stacked layers, each one can be used for feature representation. To choose the best strategy, we tested three different extraction techniques, using: (i) the last layer, (ii) the first layer and (iii) a concatenation of the four last layers. We evaluated the performance of each approach with a LR and the generic BERT model in the train and validation sets.

In our experiments, the concatenation of the four last layers did not outperformed other representations as was the case in the experiments done by \cite{Devlin-2018}. The best results were achieved using the last layer as feature, therefore we chose this representation for BERT models.

With the feature extraction strategy set, the generic BERT model was already defined. For the specialized BERT model, we still needed to set the hyperparameters for its fine-tuning. BERT has several upgradeable hyperparameters, but we limited our search to three learning rates and kept a small batch size of 32 due to memory limitations in the available GPU, keeping other parameters default.

For each tested learning rate, we retrained the model using the Internal Procedures dataset to adapt the vocabulary of the generic model and after used the last layer to extract the features, according to the best strategy found in the previous experiment. Afterwards, we trained a LR model using the training set, with the accuracy of the validation set presented in Table \ref{table:bert_finetune}.

\begin{table}[th!]
\caption{Evaluation of specialized BERT with learning rates in the validation set. In bold the best accuracy.}
\small
\centering
\begin{tabular}{lc}
\hline
\textbf{Learning Rate}  &   \textbf{Accuracy}\\
\hline
1e-4                    &   \textbf{0.690} \\
2e-5                    &   0.677 \\
5e-5                    &   0.683 \\
\hline
\end{tabular}
\label{table:bert_finetune}
\end{table}

From the results obtained in our experiments, we defined four dense feature representations.

\begin{itemize}
\itemsep0em

\item \textbf{Generic word2vec} ($W2V_{gen}$) with parameters of \cite{Hartmann-2017} and average word embeddings representation.

\item \textbf{Specialized word2vec} ($W2V_{spec}$) with parameters of \cite{Noguti-2020} and average word embeddings representation.

\item \textbf{Generic BERT} ($BERT_{gen}$) with parameters of \cite{Souza-2019} and last layer representation.

\item \textbf{Specialized BERT} ($BERT_{spec}$) with learning rate of 1e-4 and last layer representation.

\end{itemize}

The specialized language models developed in this study are available for the community, enabling the use in other experiments in the legal area that share the terminology of our vocabulary. The original data unfortunately cannot be shared at the moment, as they need to be anonymized to protect sensitive information from MPPR records.

\subsection{Supervised Learning}

We initially applied LR, SVM, Random Forest and Gradient Boosting models using each of the four feature representations and performed a grid search with 5-fold cross-validation in the training set to optimize the classifiers' parameters. The complete list of investigated parameters and best values can be found in the supplementary material.

Each of the four mentioned models were tested adding data generated synthetically by BT, doubling the size of the training set. After obtaining the best parameters, we trained the final models in the training set and evaluated the results in the test set, with the accuracy of each model presented in Table \ref{table:supervised_results}, for the original and the augmented version.

\begin{table}[th!]
\caption{Results of supervised classifiers in the test set. In bold the best accuracy for each classifier.}
\small
\centering
\begin{tabular}{llcc}
\hline
\textbf{Classifier} &   \textbf{Feature}    &   \textbf{Accuracy } &     \textbf{Accuracy }\\
					&       				&   \textbf{Original}  &     \textbf{Augment}\\

\hline
LR              & $W2V_{gen}$    	    & 0.755        		& 0.724\\
                & $W2V_{spec}$          & \textbf{0.781}    & 0.744\\
                & $BERT_{gen}$     		&0.719        		&0.696\\
                & $BERT_{spec}$         & 0.693             & 0.685\\
\hline
SVM             & $W2V_{gen}$      		& 0.705        		& 0.706\\
                & $W2V_{spec}$          & 0.752             & \textbf{0.753}\\
                & $BERT_{gen}$     		& 0.696        		& 0.697\\
                & $BERT_{spec}$         & 0.670             & 0.670\\
\hline
RF              & $W2V_{gen}$      		& 0.567        		& 0.572\\
                & $W2V_{spec}$          & 0.639             & \textbf{0.649}\\
                & $BERT_{gen}$     		& 0.571        		& 0.581\\
                & $BERT_{spec}$         & 0.565             & 0.541\\
\hline
GB              & $W2V_{gen}$      		& 0.630        		& 0.638\\
                & $W2V_{spec}$          & 0.687             & \textbf{0.707}\\
                & $BERT_{gen}$     		& 0.603        		& 0.595\\
                & $BERT_{spec}$         & 0.597             & 0.577\\
\hline
\end{tabular}
\label{table:supervised_results}
\end{table}

We notice that the use of synthetic data outperformed in most cases the classifiers trained with the original data, except in the case of LR. Even so, LR achieved the best results of all supervised models, especially when compared with the boosted classifiers. We hypothesize that those methods may require larger datasets to succeed. With restricted labeled data, as is our case, those models performed poorly compared to LR and SVM.

Another interesting result is the performance of the different feature representations. All models presented better results by using  $W2V_{spec}$, indicating that this simpler representation fits better with less complex models while also pointing out to the great benefits from training word2vec in specific corpora, in this case, the legal vocabulary. As for the BERT features, they did not performed well, leading to a hypothesis that the use of the last layer through the average subword tokens did not lead to an adequate representation in conjunction to these simpler models.

For the BERT classifier, we added a final softmax layer on top of the other layers and fine-tuned it to predict the 50 classes. The parameter search was limited to learning rate and batch sizes due to low computational resources. Since all best results were achieved with a batch size of 32, Table \ref{table:bert_classifier} presents solely the learning rate variations.

\begin{table}[th!]
\caption{Results of BERT classification in the test set. In bold the best accuracy for each language model.}
\small
\centering
\begin{tabular}{ccc}
\hline
\textbf{Examples}  &   \textbf{$BERT_{gen}$}  &   \textbf{$BERT_{spec}$}\\
\hline
4e-5    & 0.797             & 0.799 \\ 
3e-5    & \textbf{0.801}    & 0.796 \\
2e-5    & 0.795             & \textbf{0.800} \\
\hline
\end{tabular}
\label{table:bert_classifier}
\end{table}

We note that the BERT classifier reached the best results in our supervised tests. $BERT_{gen}$ slightly outperformed our specialized version, demonstrating no improvement with the fine-tuning process. Some reasons for this behavior could be the use of an insufficient amount of specialized corpora and the reduced hyperparameter investigation in consequence of computational resources restrictions.

\subsubsection{UDA}

We tested UDA with both $BERT_{gen}$ and $BERT_{spec}$, and investigated the TSA parameter, the train and validation batch sizes, and data augmentation strategies. We kept the sequence length fixed to 128 and the learning rate of 2e-5 due to prohibitive computational costs. Table \ref{table:uda} presents the results for the accuracy in the test set.

\begin{table}[th!]
\caption{Results of UDA in the test set. In bold the best accuracy for each language model.}
\small
\centering
\begin{tabular}{llcc}
\hline
\textbf{TSA} &   \textbf{Data Augment}  &   \textbf{Accuracy } 	  &  \textbf{Accuracy}\\
			 &   			   			&   \textbf{$BERT_{gen}$} &  \textbf{$BERT_{spec}$}\\

\hline
Linear  & TF-IDF      & 0.799         & 0.803 \\
                         & BT          & \textbf{0.803}& 0.805 \\
\hline
Exp     & TF-IDF      & 0.796         & 0.796 \\
                         & BT          & 0.797         & 0.801 \\
\hline
Log     & TF-IDF      & 0.794         & 0.803 \\
                         & BT          & \textbf{0.803}& \textbf{0.807} \\
\hline
\end{tabular}
\label{table:uda}
\end{table}

UDA best results showed an improvement compared to all previous tested models, especially with the linear and logarithmic TSA. The exp schedule did not work well with our data, underperforming in all tested cases. Unlike supervised BERT classifier, better performance was observed with the use of $BERT_{spec}$ in all tested cases. We also noticed an increase in accuracy with the use of BT when compared to TF-IDF replacement.

Overall, the best performance was achieved using the log TSA along with $BERT_{spec}$ and BT, with an accuracy of 80.7\%. This was also the best performance in all experiments conducted in this research, demonstrating the strength of the combined use transfer learning and data augmentation in one strategy.

\begin{figure}[ht!]
\vspace{.3in}
\centering
\includegraphics[width=8cm, trim = 0.1cm 10cm 3cm 0cm, clip]{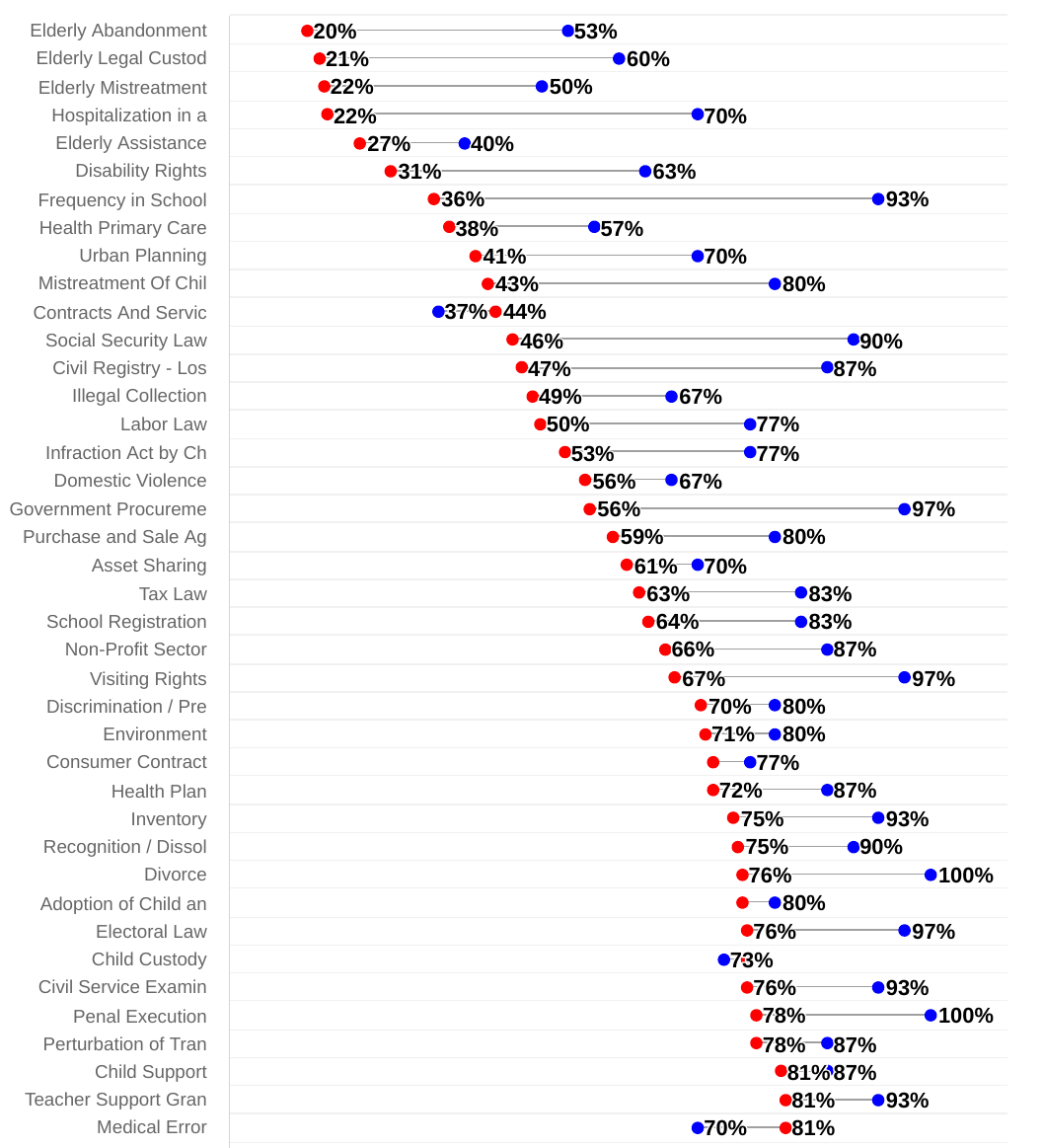}
\vspace{.3in}
\caption{Dumbbell chart of comparative performance between humans and the classifier.}
\label{fig:dumbbell_chart}
\end{figure}

As this was the best model of our experiments, we conducted a more in-depth analysis of the results. 
To analyze the individual performance of the 50 classes predicted by the model, we compared the accuracy of the classifier with the results obtained in the data audition. We considered the results as a ground truth, measuring the performance of the legal clerks in each category, and directly compared with the performance of the model. In Figure \ref{fig:dumbbell_chart}, we illustrate the accuracy achieved by humans with the red marker and UDA accuracy for the respective categories in blue. We can easily visualize that, when the red marker is to the left of the blue marker, the model's accuracy is higher than the real accuracy and vice versa. For further detailing of the classes' performances, our supplementary material describes the individual precision, recall and F1-score.

From the chart, we notice that only six categories had a higher rate of correct labels in human annotation than in the model, which are related to classes with increased assertiveness in real life. We also verify that, despite the underwhelming low values for certain classes, there is still a great improvement compared to the real-life performance, which indicates the utility of the classifier even with relatively low statistics.

Finally, we calculated the global metrics of F1-score, precision and recall for the model, as well as the accuracy @3 and @5, displayed in Table \ref{table:metrics_best_model}. We can see a general balance between precision and recall, as well as a great improvement when considering at least the three main predictions of the model, achieving an accuracy of 92\%.

\begin{table}[th!]
\caption{Global metrics for the best UDA classifier.}
\small
\centering
\begin{tabular}{ccccc}
\hline
\textbf{F1-score}   & \textbf{Precision}    & \textbf{Recall}   & \textbf{Acc @3}   & \textbf{Acc @5} \\
\hline
0.805               & 0.809                 & 0.807             & 0.921             & 0.944 \\
\hline
\end{tabular}
\label{table:metrics_best_model}
\end{table}

\section{Discussion and Conclusions}
\label{sec:conclusion}

This research sought to gather some of the most recent techniques in NLP associated with optimization in small datasets, namely the impacts of transfer learning through language models and data augmentation processes. In the addressed problem, we faced the challenge of working with a small labeled dataset and lots of unlabeled data, and investigated some approaches to improve the results in the supervised partition by leveraging the implicit knowledge of the unannotated dataset. Additionally, our study dealt with an application in the legal area, suitable for domain adaptation of language models.

Currently, due to the data protection policy, the sharing of our datasets is not allowed, since the texts contain sensitive personal information about the population served, hindering the reproducibility of results by the community. However, the fine-tuned model is available\footnote{\href{https://huggingface.co/mynoguti/BERTimbau_Legal}{https://huggingface.co/mynoguti/BERTimbau\_Legal}}.

For textual representation, we tested the context-agnostic representation word2vec and the context-dependent (Transformer-based) representation BERT. They were applied to linear and ensembles classifiers, with better performance using specialized word2vec in conjunction with LR. The BERT architecture was also used with an output layer for classification and achieved better results with the generic version, outperforming previous supervised models.


The final experiment was a combination of all previously tested techniques through the UDA algorithm. It made use of transfer learning and data augmentation and was able to attain 80.7\% of accuracy, a much higher result when compared to the manually annotated data itself, which presented an error rate of 46.7\%. When considering the @3 predictions, our model showed an increase of more than 10\% in accuracy, reaching 92\%.

We point out that the application of the model can be completely automated and integrated into MPPR's system through an API, with real-time predictions. With this workflow, it will be possible for the institution obtain a considerable gain in automating the classifications through our model, enabling its implementation with immediate gains in the quality, consistency and standardization of the records as well as allowing the allocation of MPPR contributors in more complex tasks and reverting in concrete benefits to the population as a whole.

Future works are intended to deepen the present study, in order to investigate novel language model techniques and text augmentation methods, as well as obtain more computational resources to expand our search in selected hyperparameters.

%
%
%
%

\bibliography{main}

\begin{thebibliography}{10}
\providecommand{\url}[1]{#1}
\csname url@samestyle\endcsname
\providecommand{\newblock}{\relax}
\providecommand{\bibinfo}[2]{#2}
\providecommand{\BIBentrySTDinterwordspacing}{\spaceskip=0pt\relax}
\providecommand{\BIBentryALTinterwordstretchfactor}{4}
\providecommand{\BIBentryALTinterwordspacing}{\spaceskip=\fontdimen2\font plus
\BIBentryALTinterwordstretchfactor\fontdimen3\font minus \fontdimen4\font\relax}
\providecommand{\BIBforeignlanguage}[2]{{%
\expandafter\ifx\csname l@#1\endcsname\relax
\typeout{** WARNING: IEEEtran.bst: No hyphenation pattern has been}%
\typeout{** loaded for the language `#1'. Using the pattern for}%
\typeout{** the default language instead.}%
\else
\language=\csname l@#1\endcsname
\fi
#2}}
\providecommand{\BIBdecl}{\relax}
\BIBdecl

\bibitem{CNJ-2020}
{Conselho Nacional de Justiça}, ``Justiça em números 2020: ano-base 2019,'' \url{https://www.cnj.jus.br/wp-content/uploads/2020/08/WEB-V3-Justi\%C3\%A7a-em-N\%C3\%BAmeros-2020-atualizado-em-25-08-2020.pdf}, 2020, accessed in 2020/08/27.

\bibitem{Sulea-2017}
\BIBentryALTinterwordspacing
O.-M. Sulea, M.~Zampieri, S.~Malmasi, M.~Vela, L.~P. Dinu, and J.~van Genabith, ``Exploring the use of text classification in the legal domain,'' in \emph{Proceedings of 2nd Workshop on Automated Semantic Analysis of Information in Legal Texts (ASAIL)}, London, United Kingdom, jun 2017. [Online]. Available: \url{https://arxiv.org/abs/1710.09306}
\BIBentrySTDinterwordspacing

\bibitem{Undavia-2018}
\BIBentryALTinterwordspacing
S.~Undavia, A.~Meyers, and J.~E. Ortega, ``A comparative study of classifying legal documents with neural networks,'' in \emph{2018 Federated Conference on Computer Science and Information Systems (FedCSIS)}, 2018, pp. 515--522. [Online]. Available: \url{https://ieeexplore.ieee.org/document/8511194}
\BIBentrySTDinterwordspacing

\bibitem{Silva-2018}
N.~Silva, F.~Braz, and T.~de~Campos, ``Document type classification for {B}razil’s supreme court using a convolutional neural network,'' in \emph{The Tenth International Conference on Forensic Computer Science and Cyber Law-ICoFCS,}, 10 2018, pp. 7--11.

\bibitem{Aguiar-2021}
A.~Aguiar, R.~Silveira, P.~Vladia, V.~Furtado, and J.~Neto, ``Text classification in legal documents extracted from lawsuits in brazilian courts,'' in \emph{Brazilian Conference on Intelligent Systems}, 11 2021, pp. 586--600.

\bibitem{Soh-2019}
\BIBentryALTinterwordspacing
J.~Soh, H.~K. Lim, and I.~E. Chai, ``Legal area classification: A comparative study of text classifiers on {S}ingapore supreme court judgments,'' in \emph{Proceedings of the Natural Legal Language Processing Workshop 2019}.\hskip 1em plus 0.5em minus 0.4em\relax Association for Computational Linguistics, jun 2019, pp. 67--77. [Online]. Available: \url{https://www.aclweb.org/anthology/W19-2208}
\BIBentrySTDinterwordspacing

\bibitem{Clavie-2021}
\BIBentryALTinterwordspacing
B.~Clavi{\'{e}}, A.~Gheewala, P.~Briton, M.~Alphonsus, R.~Laabiyad, and F.~Piccoli, ``Legalmfit: Efficient short legal text classification with {LSTM} language model pre-training,'' \emph{CoRR}, vol. abs/2109.00993, 2021. [Online]. Available: \url{https://arxiv.org/abs/2109.00993}
\BIBentrySTDinterwordspacing

\bibitem{Elwany-2019}
\BIBentryALTinterwordspacing
E.~Elwany, D.~Moore, and G.~Oberoi, ``{BERT} goes to law school: Quantifying the competitive advantage of access to large legal corpora in contract understanding,'' \emph{CoRR}, vol. abs/1911.00473, 2019. [Online]. Available: \url{http://arxiv.org/abs/1911.00473}
\BIBentrySTDinterwordspacing

\bibitem{Tai-2019}
\BIBentryALTinterwordspacing
C.~M.~Y. Tai, ``Effects of inserting domain vocabulary and fine-tuning {BERT} for german legal language,'' Master's thesis, Faculty of Electrical Engineering, Mathematics and Computer Science - University of Twente, Enschede - Netherlands, November 2019. [Online]. Available: \url{https://essay.utwente.nl/80128/1/Yeung_InteractionTechnology_EEMCS.pdf}
\BIBentrySTDinterwordspacing

\bibitem{Chalkidis-2020}
\BIBentryALTinterwordspacing
I.~Chalkidis, M.~Fergadiotis, P.~Malakasiotis, N.~Aletras, and I.~Androutsopoulos, ``{LEGAL}-{BERT}: The muppets straight out of law school,'' in \emph{Findings of the Association for Computational Linguistics: EMNLP 2020}.\hskip 1em plus 0.5em minus 0.4em\relax Online: Association for Computational Linguistics, nov 2020, pp. 2898--2904. [Online]. Available: \url{https://aclanthology.org/2020.findings-emnlp.261}
\BIBentrySTDinterwordspacing

\bibitem{Clavie-2021b}
\BIBentryALTinterwordspacing
B.~Clavi{\'{e}} and M.~Alphonsus, ``The unreasonable effectiveness of the baseline: Discussing {SVM}s in legal text classification,'' \emph{CoRR}, vol. abs/2109.07234, 2021. [Online]. Available: \url{https://arxiv.org/abs/2109.07234}
\BIBentrySTDinterwordspacing

\bibitem{Beltagy-2019}
\BIBentryALTinterwordspacing
I.~Beltagy, K.~Lo, and A.~Cohan, ``{S}ci{BERT}: A pretrained language model for scientific text,'' in \emph{Proceedings of the 2019 Conference on Empirical Methods in Natural Language Processing and the 9th International Joint Conference on Natural Language Processing (EMNLP-IJCNLP)}.\hskip 1em plus 0.5em minus 0.4em\relax Hong Kong, China: Association for Computational Linguistics, nov 2019, pp. 3615--3620. [Online]. Available: \url{https://aclanthology.org/D19-1371}
\BIBentrySTDinterwordspacing

\bibitem{Lee-2019}
\BIBentryALTinterwordspacing
J.~Lee, W.~Yoon, S.~Kim, D.~Kim, S.~Kim, C.~H. So, and J.~Kang, ``Bio{BERT}: a pre-trained biomedical language representation model for biomedical text mining,'' \emph{CoRR}, vol. abs/1901.08746, 2019. [Online]. Available: \url{http://arxiv.org/abs/1901.08746}
\BIBentrySTDinterwordspacing

\bibitem{Mikolov-2013}
\BIBentryALTinterwordspacing
T.~Mikolov, K.~Chen, G.~Corrado, and J.~Dean, ``Efficient estimation of word representations in vector space,'' \emph{Proceedings of Workshop at ICLR}, vol. 2013, 01 2013. [Online]. Available: \url{https://arxiv.org/abs/1301.3781}
\BIBentrySTDinterwordspacing

\bibitem{Devlin-2018}
\BIBentryALTinterwordspacing
J.~Devlin, M.-W. Chang, K.~Lee, and K.~Toutanova, ``{BERT}: Pre-training of deep bidirectional transformers for language understanding,'' in \emph{Proceedings of the 2019 Conference of the North {A}merican Chapter of the Association for Computational Linguistics: Human Language Technologies, Volume 1 (Long and Short Papers)}.\hskip 1em plus 0.5em minus 0.4em\relax Minneapolis, Minnesota: Association for Computational Linguistics, jun 2019, pp. 4171--4186. [Online]. Available: \url{https://aclanthology.org/N19-1423}
\BIBentrySTDinterwordspacing

\bibitem{Xie-2019}
\BIBentryALTinterwordspacing
Q.~Xie, Z.~Dai, E.~H. Hovy, M.~Luong, and Q.~V. Le, ``Unsupervised data augmentation for consistency training,'' \emph{CoRR}, vol. abs/1904.12848, 2019. [Online]. Available: \url{http://arxiv.org/abs/1904.12848}
\BIBentrySTDinterwordspacing

\bibitem{Wu-2016}
\BIBentryALTinterwordspacing
Y.~Wu, M.~Schuster, Z.~Chen, Q.~V. Le, M.~Norouzi, W.~Macherey, M.~Krikun, Y.~Cao, Q.~Gao, K.~Macherey, J.~Klingner, A.~Shah, M.~Johnson, X.~Liu, L.~Kaiser, S.~Gouws, Y.~Kato, T.~Kudo, H.~Kazawa, K.~Stevens, G.~Kurian, N.~Patil, W.~Wang, C.~Young, J.~Smith, J.~Riesa, A.~Rudnick, O.~Vinyals, G.~Corrado, M.~Hughes, and J.~Dean, ``Google's neural machine translation system: Bridging the gap between human and machine translation,'' \emph{CoRR}, vol. abs/1609.08144, 2016. [Online]. Available: \url{http://arxiv.org/abs/1609.08144}
\BIBentrySTDinterwordspacing

\bibitem{Hartmann-2017}
\BIBentryALTinterwordspacing
N.~S. Hartmann, E.~Fonseca, C.~D. Shulby, M.~V. Treviso, J.~S. Rodrigues, and S.~M. Aluísio, ``Portuguese word embeddings: Evaluating on word analogies and natural language tasks,'' in \emph{Proceedings of Symposium in Information and Human Language Technology}.\hskip 1em plus 0.5em minus 0.4em\relax Sociedade Brasileira de Computação, 2017. [Online]. Available: \url{https://sol.sbc.org.br/index.php/stil/article/view/4008}
\BIBentrySTDinterwordspacing

\bibitem{Noguti-2020}
\BIBentryALTinterwordspacing
M.~Y. {Noguti}, E.~{Vellasques}, and L.~S. {Oliveira}, ``Legal document classification: An application to law area prediction of petitions to public prosecution service,'' in \emph{2020 International Joint Conference on Neural Networks (IJCNN)}, 2020, pp. 1--8. [Online]. Available: \url{https://ieeexplore.ieee.org/abstract/document/9207211}
\BIBentrySTDinterwordspacing

\bibitem{Souza-2019}
\BIBentryALTinterwordspacing
F.~Souza, R.~Nogueira, and R.~Lotufo, ``Portuguese named entity recognition using {BERT-CRF},'' \emph{arXiv preprint arXiv:1909.10649}, 2019. [Online]. Available: \url{http://arxiv.org/abs/1909.10649}
\BIBentrySTDinterwordspacing

\bibitem{Wagner-2018}
\BIBentryALTinterwordspacing
J.~Wagner, R.~Wilkens, M.~Idiart, and A.~Villavicencio, ``The {brWaC} corpus: A new open resource for brazilian portuguese,'' in \emph{Proceedings of the Eleventh International Conference on Language Resources and Evaluation}, 05 2018. [Online]. Available: \url{https://aclanthology.org/L18-1686/}
\BIBentrySTDinterwordspacing

\end{thebibliography}
\bibliographystyle{IEEEtran}

\end{document}